\documentclass[10pt, a4paper]{article}
\usepackage[thai,english]{babel}
\usepackage[utf8x]{inputenc}
\usepackage{fonts-tlwg}

\usepackage{lrec}
\usepackage{multibib}
\newcites{languageresource}{Language Resources}
\usepackage{graphicx}
\usepackage{subfigure}

\usepackage{tabularx}
\usepackage{soul}

\usepackage{hyperref}
\usepackage{xstring}

\usepackage{color}

\newcommand{\secref}[1]{{\StrSubstitute{\getrefnumber{#1}}{.}{}}}

\usepackage{todonotes}

\usepackage{xargs}
\usepackage{amsmath}

\let \citeauthor \newcite

\usepackage{colortbl} 
\usepackage{xspace}

\usepackage{tikz}
\usetikzlibrary{arrows,shapes}

\usepackage{enumitem} 

\usetikzlibrary{calc}
\usetikzlibrary{decorations.pathmorphing}

\newlength\LineWidth
\newlength\Amplitude
\newlength\SegLength

\definecolor{HLcolor}{RGB}{240,0,0}

\newcommand\tikzmark[1]{%
  \tikz[overlay,remember picture] \node (#1) {};}

\makeatletter
\newcommand{\highlight@DoHighlight}{
  \draw[HLcolor,line width=\LineWidth,decorate,decoration={zigzag,amplitude=\Amplitude,segment length=\SegLength}]  ($(begin highlight)+(0,-2pt)$) -- ($(end highlight)+(0,-2pt)$) ;
}

\newcommand{\highlight@BeginHighlight}{
  \coordinate (begin highlight) at (0,0) ;
}

\newcommand{\highlight@EndHighlight}{
  \coordinate (end highlight) at (0,0) ;
}

\newdimen\highlight@previous
\newdimen\highlight@current

\DeclareRobustCommand*\highlight[1][]{%
  \SOUL@setup
  \def\SOUL@preamble{%
    \begin{tikzpicture}[overlay, remember picture]
      \highlight@BeginHighlight
      \highlight@EndHighlight
    \end{tikzpicture}%
  }%
  \def\SOUL@postamble{%
    \begin{tikzpicture}[overlay, remember picture]
      \highlight@EndHighlight
      \highlight@DoHighlight
    \end{tikzpicture}%
  }%
  \def\SOUL@everyhyphen{%
    \discretionary{%
      \SOUL@setkern\SOUL@hyphkern
      \SOUL@sethyphenchar
      \tikz[overlay, remember picture] \highlight@EndHighlight ;%
    }{%
    }{%
      \SOUL@setkern\SOUL@charkern
    }%
  }%
  \def\SOUL@everyexhyphen##1{%
    \SOUL@setkern\SOUL@hyphkern
    \hbox{##1}%
    \discretionary{%
      \tikz[overlay, remember picture] \highlight@EndHighlight ;%
    }{%
    }{%
      \SOUL@setkern\SOUL@charkern
    }%
  }%
  \def\SOUL@everysyllable{%
    \begin{tikzpicture}[overlay, remember picture]
      \path let \p0 = (begin highlight), \p1 = (0,0) in \pgfextra
        \global\highlight@previous=\y0
        \global\highlight@current =\y1
      \endpgfextra (0,0) ;
      \ifdim\highlight@current < \highlight@previous
        \highlight@DoHighlight
        \highlight@BeginHighlight
      \fi
    \end{tikzpicture}%
    \the\SOUL@syllable
    \tikz[overlay, remember picture] \highlight@EndHighlight ;%
  }%
  \SOUL@
}
\makeatother

\newcommand{\MarkText}[1]{
    \colorlet{HLcolor}{red}
    \setlength\Amplitude{1pt}%
    \setlength\SegLength{5pt}%
    \tikzmark{endquote}\tikzmark{beginquote}\highlight{#1}%
}

\title{\MarkText{Misspelling} Semantics in Thai}

\name{Pakawat Nakwijit$^{1}$ and Matthew Purver$^{1,2}$}

\address{\quad\quad\quad\quad$^{1}$Cognitive Science Research Group \quad\quad\quad\quad\quad\quad $^{2}$Department of Knowledge Technologies\\
\hspace*{-1cm}School of Electronic Engineering and Computer Science \quad\quad\quad\quad Jožef Stefan Institute \\
Queen Mary University of London, UK \quad\quad\quad\quad\quad\quad\quad\quad Ljubljana, Slovenia\\
\{p.nakwijit,  m.purver\}@qmul.ac.uk\\}

\abstract{
User-generated content is full of misspellings. Rather than being just random noise, we hypothesise that many misspellings contain hidden semantics that can be leveraged for language understanding tasks. This paper presents a fine-grained annotated corpus of misspelling in Thai, together with an analysis of misspelling intention and its possible semantics to get a better understanding of the misspelling patterns observed in the corpus. In addition, we introduce two approaches to incorporate the semantics of misspelling: Misspelling Average Embedding (MAE) and Misspelling Semantic Tokens (MST). Experiments on a sentiment analysis task confirm our overall hypothesis: additional semantics from misspelling can boost the micro F1 score up to 0.4-2\%, while blindly normalising misspelling is harmful and suboptimal. \\ \newline \Keywords{Thai, misspelling, text normalisation, word embedding} }

\begin{document}

\maketitleabstract

\section{Introduction}
The idea that feelings and emotions can be expressed and shared with others through text is now familiar  \cite{alsayat2021improving}. Conventionally, punctuation and typographic styling (italic, bold, and underlined text) are used as prosodic indicators to emphasise an important word. However, with the fast and widespread internet adoption, the communication medium now is not limited to formal written texts such as newspapers and books. The daily conversation appears everywhere on the internet leading to a new orthographic style much closer to the spoken form: informal, context-dependent and, importantly, full of misspellins \footnote{Some typographical and grammatical misspellings are intentionally kept intact in the paper by authors} \cite{brody-diakopoulos-2011-cooooooooooooooollllllllllllll}.

In English, more than 70\% of documents on the internet contain some form of misspelling \cite{ringlstetter2006orthographic}. Misspelling sometimes occurs unintentionally when people hit two adjacent keys on the keyboard in a single keystroke, accidentally add/miss letters when they type, or due to a lower level of language proficiency. However, a large percentage of misspelling is intentional. Intentionally misspelt words can be used as prosody to provide additional clues about the writer's attitude. They can be used to show affection towards an interlocutor, emphasise the sentiment of a word, avoid offensive meaning or even represent the speaker's identity \cite{brody-diakopoulos-2011-cooooooooooooooollllllllllllll, tavosanis2007causal, contextual-bearing}.

However, this misspelling semantics has been largely ignored in the literature. Many previously published studies are limited to formal and well-curated corpora such as Wikipedia to avoid misspelling noise which is likely to interfere with the model accuracy \cite{devlin2018bert, grave2018learningfasttext, sun2020advbert}. In studies that focus on informal text, one standard practice is to ignore misspelling, effectively treating misspelled tokens as distinct from their standard equivalents. Another is lexical normalization before training: transforming non-standard tokens into a more standardised form to reduce the number of out-of-vocabulary tokens \cite{haruechaiyasak-kongthon-2013-lextoplus, cook-stevenson-2009-unsupervised, han-baldwin-2011-lexical, liu-etal-2012-broad}. Both approaches therefore ignore the hidden semantics of misspelling, either by explicitly removing it or by losing the connection to the standard form.

In this paper, we instead propose that misspelling should not be discarded or ignored. The hidden semantics within misspelling tokens can provide useful information that can be extracted to comprehend the sentiment of a sentence.

Moreover, much of the research up to now has been done only on English texts. Its findings are potentially missing out on valuable information that can be useful for generalisation to other languages, particularly those in which misspelling phenomena may be even more complex and meaningful. In this paper, we focus on Thai. Thai is under-studied despite its unique linguistic features that are vastly different to the English speaking world: for example, the use of tone marker and vowel duration in Thai leads to a variety of ways to form a word and various strategies to misspell it to convey additional meaning. We also suspect that because Thai is an analytic language, less information could be expressed syntactically, with more reflected directly on the surface form of a word. 

In this paper, our aim is to raise awareness of the importance of the semantics of misspelling. We present a new fine-grained annotated corpus of misspelling in Thai and demonstrate two approaches that can be used to incorporate the misspelling semantics to state-of-the-art sentiment analysis classifiers.

\section{Related Works}
Misspellings over the internet have been studied since the early 2000s. \citeauthor{ringlstetter2006orthographic} investigated and classified various types of orthographic errors, including typing errors, spelling errors, encoding errors and OCR errors. Error detection was developed to normalize the web corpus. \citeauthor{tavosanis2007causal} presented a similar classification, but recognized intentional deviations as a different class of misspelling. However, with emerging of new technology, these categories are now outdated. Encoding and OCR errors are not prevalent in the current internet corpus. In addition, intentional misspelling could be more than a stylistic choice to overcome technical limitations or circumvent automatic indexing or censoring mechanisms. In this paper, we propose novel classification criteria that suit modern social text corpus, including unintentional and intentional misspelling and present a fine-grained analysis of misspelling patterns observed in our corpus.

More recent works started investigating different types of misspelling formation. \citeauthor{cook-stevenson-2009-unsupervised} and  \citeauthor{han-baldwin-2011-lexical} presented a consistent observation that the majority of the misspelling found on the internet is from morphophonemic variations (transformation of surface form of a word but conserve similar pronunciation) and abbreviations. This finding is then used as a guideline to build their lexical normalization models. \citeauthor{liu-etal-2012-broad} extended previous normalization approaches by incorporating a phenomenon called ``visual priming'' (a phenomenon when a misspelling token can be recognized based on a commonly used word). These three studies suggested that misspelling is not arbitrary. It associates with human cognition and perception of a language. However, they utilized misspelling information only in the lexical normalization, discarding all misspelling terms during model training. In contrast, this paper argues that the misspelling tokens should not be normalized and discarded. We also present two approaches to leverage them and show a noticeable improvement on the sentiment analysis task. 

It was pointed out by \citeauthor{al-sharou-etal-2021-towards} that textual noise is not always harmful to the system. It could carry a meaning that is important for a certain task. \citeauthor{brody-diakopoulos-2011-cooooooooooooooollllllllllllll} showed that repetitive characters in text are closely related to subjective words. They also suggested that it might associate with prosodic indicators, which are commonly used in verbal communication. 

\citeauthor{john2019context} suggested that including character repetition and word capitalization to a sentiment classification model gain a substantial improvement. These studies support our hypothesis that misspelling has inherent semantics that correlates with the sentiment of a sentence. However, the studies are limited to traditional machine learning models. In contrast, we present new approaches that are suitable for SOTA neural models, both shallow neural networks and deeper models such as BERT. We evaluated our results with 2 models: LSTM with static fastText embeddings \cite{grave2018learningfasttext} and a pre-trained BERT-like model: WangchanBERTa \cite{lowphansirikul2021wangchanberta}.

Although extensive research has been carried out on misspellings in English, few studies exist on other languages. In this paper, we study misspellings in Thai as it has different orthography and phonology to English, and thus may provide insights not yet considered in the literature. One early work on misspelling in Thai is proposed by \citeauthor{haruechaiyasak-kongthon-2013-lextoplus}. They identified four intentional misspellings classes: insertion (character repetitive), transformation (homophonic and syllable trimming), transliteration (foreign words written in Thai), and onomatopoeia (words that resemble the non-verbal sound). However, in the paper, their model could only detect repeated characters. \citeauthor{poolsukkho2018thainorm} extended it by employing IPA similarity to the existing model to include homophonic words. The main limitation of their model, however, is the low coverage as they used a dictionary-based model. \citeauthor{lertpiya2020thaispellingcorreciton} addressed the coverage issue by developing neural-based models on a larger corpus. Their model used two separated models; misspelling detection and misspelling correction. It significantly improved the earlier works. Similar to previous works in English, these studies only focused on normalizing texts and discarding misspelling information. 
 
\section{Misspelling Corpus}
In this section, we present a new fine-grained Thai misspelling corpus. It is an extension of the Wisesight Sentiment corpus \cite{arthit-2019-wisesight}. It is widely used as a standard benchmark for Thai sentiment analysis. The data were collected from various social media in Thailand from 2016 to early 2019. It consists of posts, comments, informal conversations, news headlines and advertisements. Each message was annotated into three categories: positive, neutral, and negative.\footnote{Originally, wisesight sentiment corpus has four classes, including a \textit{question} class. However, based on our observation, its annotation description is self-contradictory, resulting in significantly inconsistent labelling. It also has relatively little data, so we decided to ignore and treat it as neutral to reduce the complexity of the task.} Train, validation and test datasets are provided, consisting of 21628, 2404 and 2671 sentences. 

Our new corpus is based on a sample of 3000 sentences from the training data. It is manually annotated by five recruited annotators. They are Thai native speakers to ensure that they can fully comprehend the sentiment of the given sentences. We employed a two-iterative annotation strategy where the annotators were asked to label misspellings according to our guideline. We then evaluated 100 samples and gave feedback to the annotators before asking them to re-label the data again. Each sentence was guaranteed to be annotated by three annotators. Each misspelling was labelled as \textit{intentional} or \textit{unintentional} based on the criteria described in Section~\secref{sec_misp_semantics}.

In total, we collected 1484 misspelling words with 728 unique token types. There are 971 sentences that have at least one misspelling. They account for 32.4\% of the annotated training data. Class distribution of the misspelling sentences is 39.3\%, 35.6\% and 25.1\% for negative, positive and neutral, respectively. 

\begin{figure}[h]
    \centering
    \includegraphics[width=0.45\textwidth]{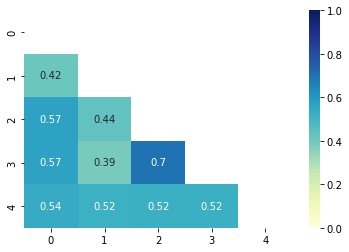}
    
    \caption{Inter-Annotator Agreement on misspelling intention among annotators for annotators 0-4 }
    \label{fig:iaa}
    
\end{figure}

We used Cohen's kappa \cite{artstein2008IAA} to visualise inter-annotator agreement among annotators on the intention class of a misspelt word: see Figure~\ref{fig:iaa}. Results show that classifying misspelling intention might not be as trivial as expected, but it still contains a moderate agreement level. 

\begin{figure}[h]
    \centering
    \includegraphics[width=0.45\textwidth]{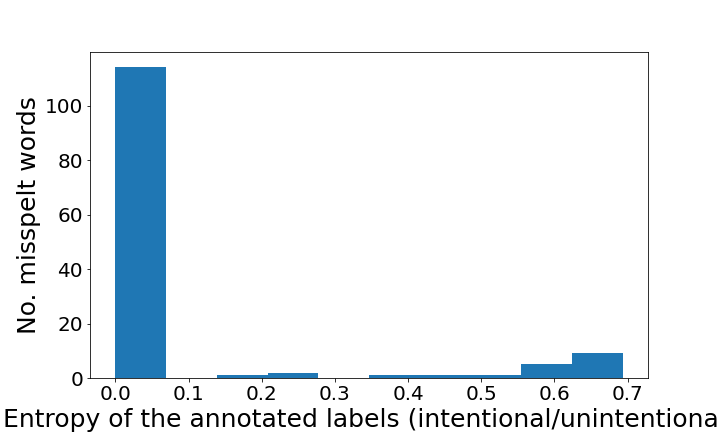}
    \caption{Entropy of the label from each misspelt word (considering only misspelt words observed more than 5 times)}
    \label{fig:int_entropy}
    
\end{figure}

\begin{figure}[ht]
    \centering
    \includegraphics[width=0.45\textwidth]{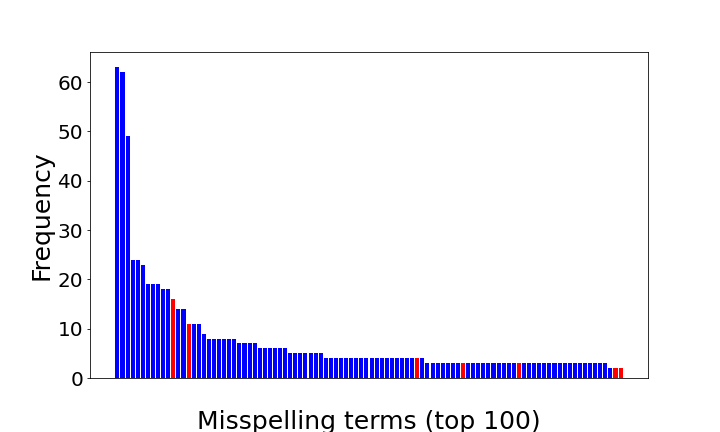}
    \caption{Term frequency of the misspelling (top 100); intentional (blue) and unintentional (red)}
    \label{fig:tf}
    
\end{figure}

In the end, we got 572 and 156 for \textit{intentional} and \textit{unintentional} misspelling terms (unique token type). The frequency distribution shows that most misspelling terms are intentional (up to 86.4\%). Only two unintentional words were observed in the top 20 most frequent words (see Figure~\ref{fig:tf}). Unsurprisingly, we observed that intentional words are mostly sentiment-related words. On the other hand, unintentional words are mainly sentence-final particles and typos which play little or no role in the sentiment (see Table~\ref{tab:freqterm}). In addition, we calculated the entropy of the label from each misspelt term; see Figure~\ref{fig:int_entropy}. It shows that intention of misspelt words is strongly consistent, confirming that the misspelling is not arbitrary. 

\begin{table}
    \centering
        \begin{tabular}{|c|c|}
              \hline
              Intentional &Unintentional\\
              \hline
                \thaitext{แม่ง} &
                \thaitext{ค่ะ} \\
              \hline
                \thaitext{คับ} &
                \thaitext{คะ} \\
              \hline
                \thaitext{กุ} &
                \thaitext{จ่ะ} \\
              \hline
                \thaitext{สัส} &
                \thaitext{แล้ว} \\
              \hline
                \thaitext{มากกก}  &
                \thaitext{อ้ะ} \\
              \hline
        \end{tabular}
    
    \caption{Top 5 most frequent misspelling terms}
    \label{tab:freqterm}
\end{table}

In addition, we annotated the Wisesight test set with intention classes similar to the training data. Only one annotator was asked to correct the misspelling and categorised it into 10 classes according to how it was misspelt (details given in Section~\secref{sec_misp_semantics} below). Label distribution is shown in Table~\ref{tab:pattern}.  Even though our study focuses on the semantic functions of the misspelling, it is non-trivial to classify them directly as they are paralinguistic and vague. We instead consider that investigating how people misspell a word could give a more definitive answer.

\begin{table}
    \centering
        \begin{tabular}{|l|c|}
            \hline
            Misspelling Pattern & Count\\
            \hline
            \hline
                Character repetition &
                300 \\
            \hline
                Vowel substitutions &
                428 \\
            \hline
                Tone modification &
                402 \\
            \hline
                Consonant deviation &
                51 \\
            \hline
                Others &
                371 \\
            \hline
            \hline
                Simplifying &
                217 \\
            \hline
                Ad hoc abbreviation &
                1 \\
            
            \hline
            \hline
                Tone confusion &
                211 \\

            \hline
                Consonant confusion &
                15 \\
            \hline
                Typos &
                106 \\
            
            \hline
        \end{tabular}
    \caption{Misspelling Patterns from the Wisesight test set}
    \label{tab:pattern}
\end{table}

Lastly, because tokenisation plays a vital role in a downstream task   \cite{beaufort-etal-2010-hybrid, alkaoud-syed-2020-importance}, mistokenising a sentence can drastically change its meaning. This effect will be more extreme on a corpus with more misspellings and when the language of interest has ambiguous word boundaries, such as Thai. To control the impact of tokenisation, DeepCut \cite{Kittinaradorn2019deepcut} was used to pre-tokenise all sentences in the corpus after the annotating process. In the test set, our custom post-process was applied to ensure the number of tokens from the original sentences and the corrected sentences matches.

\section{Misspelling Semantics}
\label{sec_misp_semantics}
In this section, we discuss misspelling patterns observed in the corpus and its possible semantics.

The term ``misspelling'' has been generally defined as ``when a word is spelt in a way that deviates from reference dictionaries, standardized or accepted norms or recognized usage" \cite{al-sharou-etal-2021-towards}. It includes typos, ad hoc abbreviations, unconventional spellings, phonetic substitutions and lexical deviation. In this paper, we introduce a new term, ``misspelling semantics'', to consider the meaning behind how and why people misspellingly form a word. 

We use common spelling based on Google search auto-correction and Thai Royal Institute Dictionary as the reference dictionaries \cite{thaidictionary}. In contrast to \citeauthor{haruechaiyasak-kongthon-2013-lextoplus}, we do not consider transliterated forms as a misspelling as there is no standard transliterated spelling, so it is difficult to decide whether a word is a misspelling. In addition, we also ignore misspellings due to misuse of Thai orthographic signs such as ``\thaitext{ๆ}'' and ``\thaitext{ฯ}'' (introduced by \cite{limkonchotiwat-etal-2021-handling}). As it typically is a stylistic error, it is not related to semantics.

In the initial data exploration, we interviewed Thai natives to get opinions on the common misspelling patterns used in daily conversation. We asked them to classify each misspelt word into two classes: unintentional and intentional. The criteria were formalised into a series of 3 questions. 

\textbf{1. Does it convey an additional meaning/emotion?}
We asked annotators to observe an additional meaning when a misspelt word and the original counterpart cannot be interchangeable within the same context. This additional function could be amplifying the meaning, euphemism, showing affection, friendliness or respect. 

\textbf{2. Does the misspelt word need more/less effort to type?}
How people misspell a word is closely related to a keyboard layout. According to our interview, one reason to misspell a word is because some misspelt words require less effort to type. It might be due to closer key buttons, fewer keypress or no shift key required. 

\textbf{3. Is the word not a commonly misspelt word?}
This question was asked to eliminate misspellings due to varying levels of language proficiency and accidental typographical error. Because of the complexity of Thai writing system, a variety of mistakes could be observed, but they are unintentional without useful semantics.

Answering \textit{yes} to one of these questions is considered as an \textit{intentional}. Otherwise, \textit{unintentional}. Although misspelling from the last two questions might not evidently associate with the semantics of a sentence, it should be noted for the completeness of the study. 

Based on the criteria, we observed 10 misspelling patterns found in our corpus. The patterns are grouped based on their related question. Curated examples are given together with their normalised form and English translation in [\dots] and (Eng: \dots), respectively.

\subsection{Does it convey an additional meaning/emotion?}
We observed five misspelling patterns falling under this criterion. 

\subsubsection{Character repetition}
Character repetition is the most common misspelling pattern mentioned in the literature. As suggested by \citeauthor{brody-diakopoulos-2011-cooooooooooooooollllllllllllll}, the character repetition might be a textual representation mimicking how people prolong a sound in a conversation to amplify the meaning of a word or to draw attention. 

Interestingly, character repetition in Thai can be observed both in the vowel and the final consonant of a word, unlike in English, where it is predominately found in the vowel. This might be due to grammatical differences as Thai has no inflection. Repetition in the final consonant does not interfere with the presence of a grammatical suffix, e.g. /-s/ or /-ed/. 

\begin{flushleft}
\textbf{Examples}: 
\begin{itemize}
    \setlength{\itemsep}{0.1em}
    \item{
        \thaitext{น่า\textbf{กินนนนนนนนนนนนน}}
        [\thaitext{น่า\textbf{กิน}}]\newline
        (Eng: \textit{Looks delicious})
    } 
    
    \item{
        \thaitext{\textbf{มึงงงงงงง} ไป\textbf{กันๆๆๆๆ}}
        [\thaitext{\textbf{มึง} ไป\textbf{กัน}}]\newline
        (Eng: \textit{Hey! let's do this})
    }
\end{itemize}
\end{flushleft}

\subsubsection{Vowel substitutions}
In Thai phonology, there are nine basic vowel monophthongs. Each of them is pronounced with either a short or long duration \cite{iwasaki2005referencethai}. Vowel substitutions refer to when people intentionally substitute a short vowel with its long vowel (and vice versa) to form a new word. It is the most common misspelling pattern observed in our corpus. 

The previous studies have shown there is a correlation between long vowel sounds and taste expectations of sweetness \cite{pathak2021sooo}. In our context, it can be seen as a way to de-emphasize the offensive meaning of a word. 

On the other hand, shortening vowels is not commonly done. Based on our observation, we cannot find clear sentiment-related semantics. However, we suggest that it might be a form of vowel weakening which is often found in fast speech.

\begin{flushleft}
\textbf{Examples}: 
\begin{itemize}
    \setlength\itemsep{0.1em}
    \item{
        \thaitext{\textbf{เมิง}พร้อม\textbf{กาน}\textbf{ยางงงง}}
        [\thaitext{\textbf{มึง}พร้อม\textbf{กัน}\textbf{รึยัง}}]\newline
        (Eng: \textit{Are you ready?})
    } 
    
    \item{
        \thaitext{\textbf{เส็ด}\textbf{มะไหร่} \textbf{กุ}พร้อม\textbf{ละ} }\newline
        [\thaitext{\textbf{เสร็จ}\textbf{เมื่อไหร่} \textbf{กู}พร้อม\textbf{แล้ว}}]\newline
        (Eng: \textit{Ready? I'm ready})
    }
\end{itemize}
\end{flushleft}

Another pattern of the substitutions can be observed in vowels with an ending consonant sound.\footnote{Modern Thai literature does not count them as a vowel but as a syllable. However, to minimize the complexity of the analysis, we decide to keep them as a vowel.} Because of their inherent consonant, they can be written in 2 forms; with or without a presence of the consonant grapheme; 
\thaitext{อำ},
\thaitext{ไอ}, 
\thaitext{ใอ} and 
\thaitext{อัม}/\thaitext{อรรม},
\thaitext{อัย}, 
\thaitext{อัย} respectively.
A word with these vowels can be written in either form to represent the same sound. We observe that they are not interchangeable. Transformed words could provide a feeling of informality or friendliness to a word, but this does not always correspond to the sentiment.

\begin{flushleft}
\textbf{Examples}: 
\begin{itemize}
    \setlength\itemsep{0.1em}
    \item{
        \thaitext{ขอบ\textbf{จัย}จ้า }
        [\thaitext{ขอบ\textbf{ใจ}จ้า}]\newline
        (Eng: \textit{Thank you} )
    }
    \item{
        \thaitext{ก็แค่\textbf{กำ}เก่า}
        [\thaitext{ก็แค่\textbf{กรรม}เก่า}]\newline
        (Eng: \textit{Just your old deeds})
    }
\end{itemize}
\end{flushleft}

\subsubsection{Tone modification}
Tone is a crucial component in Thai. Words that are similar but pronounced with different tones usually have different, unrelated meanings. However, in informal conversation, the use of tone is more flexible. The introduction of social media leads to a shift in written texts where people tend to express tone differently from the standard writing to reflect the actual use of the tone in speech.

The annotated data suggested that there is an increase in the use of higher tones. We suspect that the shift in tone usage in Thai might be influenced by the use of rising intonation in English. However, there is no sentiment-related meaning to be observed from this pattern.

\begin{flushleft}
\textbf{Examples}: 
\begin{itemize}
    \setlength\itemsep{0.1em}
    \item{
        \thaitext{โชคดี\textbf{น๊าคร๊าบ}}
        [\thaitext{โชคดี\textbf{นะครับ}}]\newline
        (Eng: \textit{Good luck})
    }
    \item{
        \thaitext{กินข้าวกัน\textbf{มั้ย}ครับ}
        [\thaitext{กินข้าวกัน\textbf{ไหม}ครับ}]\newline
        (Eng: \textit{Would you like to have dinner together, tomorrow?})
    }
\end{itemize}
\end{flushleft}

\subsubsection{Consonant Deviation}
Consonant deviation is about changes in the consonant of a word. It can be observed both in the initial and final positions. Based on our observations, there are no strict rules on how to transform a consonant. It could be transformed into a consonant cluster, or consonant clusters can be reduced to a single consonant. However, the consonant clusters observed are limited to clusters with the additional /w/, /l/ and /r/.

Adding a new consonant and replacing the consonant are also common. The consonant in the former case usually comes with a cancellation mark to dictate it into a silent voice. Replacing a consonant is more flexible but not arbitrary. However, we have not yet found rules to explain this linguistic phenomenon. Interestingly, a certain number of added/replaced consonants might be obtained from foreign languages. We believe that this may be used to mimic the ending sounds, such as /st/ in ``first'' and /ch/ in ``watch'', that are not spoken in Thai. 

In general, our data suggests that consonant deviation could be an indicator of friendliness and playfulness, which is likely to correspond to positive sentiment.

\begin{flushleft}
\textbf{Examples}: 
\begin{itemize}
    \setlength\itemsep{0.1em}
    \item{
        \thaitext{\textbf{เธอว์} อร่อย\textbf{มว๊าก}\textbf{เรย}\textbf{จร้า}}
        [\thaitext{\textbf{เธอ} อร่อย\textbf{มาก}\textbf{เลย}\textbf{จ้า}}]\newline
        (Eng: \textit{It's so tasty})
    }
    \item{
        \thaitext{ไม่ได้\textbf{ฮ้าบ}}
        [\thaitext{ไม่ได้\textbf{ครับ}}]\newline
        (Eng: \textit{No})
    }
    \item{
        \thaitext{\textbf{เพิ่ลๆ}\textbf{ครัช} \textbf{เสด}รึยัง}
        [\thaitext{\textbf{เพื่อนๆ}\textbf{ครับ} \textbf{เสร็จ}รึยัง}]\newline
        (Eng: \textit{Guys, are you ready?})
    }
\end{itemize}
\end{flushleft}

\subsubsection{Others}
Because of the diverse culture of internet users, new words are invented every day from the existing vocabulary. The pattern to describe how people form a word is, sometimes, more complex than changing a vowel, consonant or tone. 

In some extreme cases, a new sub-language is created to represent a specific group of people, such as LGBTQ+ or particular dialects \cite{tavosanis2007causal, contextual-bearing}. It, later, becomes a stylish identity. One example from Thai is ``Skoy language''. Its unique feature is the excessive use of high tone markers and complex characters. No single transformation can describe the language; it consists of a combination of several transformations and the context. Using these sub-languages often inherits the public image of the group into the text, such as social status, age group, and personality.


\begin{flushleft}
\textbf{Examples}: 
\begin{itemize}
    \setlength\itemsep{0.1em}
    \item{
        \thaitext{\textbf{เมพขิงๆ}เลยวะ \textbf{สุโข่ย}}
        [\thaitext{\textbf{เก่งสุดๆ}เลยวะ \textbf{สุดยอด}}]\newline
        (Eng: \textit{Genius})
    }
    \item{
        \thaitext{อ่านออก\textbf{ม๊อ๊คร๊}}?
        [\thaitext{อ่านออก\textbf{ไหมอะคะ}}?]\newline
        (Eng: \textit{Can you read this?})
    }
\end{itemize}
\end{flushleft}

In less extreme cases, we observe words where some letters were replaced with numbers or homorph glyphs; visually similar letters. Some words were changed into other words that are not semantically correct in the context. It could be considered as a stylistic choice. However, it can also be used to avoid controversial content detection from a platform such as swear words and sexual words. One example is the word ``\thaitext{เสือก} (Eng: \textit{mind your bussiness})'' , which is censored by Pantip.com -- a popular Thai webboard. To avoid the censorship of the platform, people misspelt it into ``\thaitext{เผือก} (Eng: \textit{taro})''.


However, because of the mixed patterns and their insubstantial numbers of observations, we cannot conclude how this type of misspelling accounts for the sentiment of a sentence. 

\subsection{Does the misspelt word need more/less effort to type?}
We observed two misspelling patterns related to typing. 

\subsubsection{Simplifying}
To simplify a word is to shorten a word for convenience to type or to read. It could be on a phonological level where the vowel of a word is changed into the short /a/ vowel or a syllable is completely removed. Another simplifying type is on the surface level, where a character in a word is changed to a more common character. 

\begin{flushleft}
\textbf{Examples}: 
\begin{itemize}
    \setlength\itemsep{0.1em}
    \item{
        \thaitext{\textbf{ก้มะรุ้}เหมือนกัน}
        [\thaitext{\textbf{ก็ไม่รู้}เหมือนกัน}]\newline
        (Eng: \textit{I don't know either})
    }
    \item{
        \thaitext{\textbf{ไปรสนีไท}ไม่โอเลย}
        [\thaitext{\textbf{ไปรษณีย์ไทย}ไม่โอเคเลย}]\newline
        (Eng: \textit{Thai post is not good})
    }
\end{itemize}
\end{flushleft}

\subsubsection{Ad hoc abbreviation}
Ad hoc abbreviation is a shortened form of a word or phrase that is not commonly used or requires context to understand. It could also be used to encode information known only within a small group of people.

\begin{flushleft}
\textbf{Examples}: 
\begin{itemize}
    \setlength\itemsep{0.1em}
    \item{
        \thaitext{\textbf{พน} ร้านปิดครับ}
        [\thaitext{\textbf{พรุ่งนี้} ร้านปิดครับ}]\newline
        (Eng: \textit{Tomorrow, our shop is closed})
    }
    \item{
        \thaitext{\textbf{ผนงรจตกม}}
        [\thaitext{\textbf{ผู้นำโง่เราจะตายกันหมด}}]\newline
        (Eng: \textit{A stupid leader will lead us all to die})
    }
\end{itemize}
\end{flushleft}

\subsection{Is the word not a commonly misspelt word?}
Please note that the following categories are by no means comprehensive. We presented only two common patterns observed in the corpus. 

\subsubsection{Tone confusion}
The presence of tone in Thai makes it tricky to read and write. Thai consists of five distinct tones, which are realized in the vowels, but indicated in the script by a combination of the class of the initial consonant (high, mid or low), vowel length (long or short), closing consonant (plosive or sonorant) and tone marks. Because of the complex tone system, tone confusion is prevalent in internet conversation, even among Thai people. One example is the use of \thaitext{คะ} and \thaitext{ค่ะ}. The former is often used in a question sentence, while the latter is used as a sentence-ending particle. Misinterpreting them without context often results in a completely different meaning.

\begin{flushleft}
\textbf{Examples}: 
\begin{itemize}
    \setlength\itemsep{0.1em}
    \item{
        \thaitext{แม่ค้าร้านนี้ใจดีมาก\textbf{คะ} อุดหนุนเยอะๆ\textbf{นะค่ะ}}\newline
        [\thaitext{แม่ค้าร้่านนี้ใจดีมาก\textbf{ค่ะ} อุดหนุนเยอะๆ\textbf{นะคะ}}]\newline
        (Eng: \textit{The seller of this shop is very kind. Please support her})
    }
    \item{
        \thaitext{เอา\textbf{ละ} จะกิน\textbf{ล่ะนะ}}} 
        [\thaitext{เอา\textbf{ล่ะ} จะกิน\textbf{ละนะ}}]\newline
        (Eng: \textit{Let's eat!})
\end{itemize}
\end{flushleft}

\subsubsection{Consonant confusion}
Due to geological factors, Thai is greatly influenced by Indic languages such as Sanskrit and Pali. It can be observed by a number of duplicate letters that represent separate sounds in Sanskrit and Pali but are not distinct sounds in Thai. The language has also inherited orthographical rules to conserve the etymology of a word. As a result, Thai has a complex writing system that does not have a one-to-one correspondence between phonemes and graphemes. Writing a word in Thai requires memorizing its exact spelling or its etymology.

\begin{flushleft}
\textbf{Examples}: 
\begin{itemize}
    \setlength\itemsep{0.1em}
    \item{
        \thaitext{\textbf{ปฏิมากรรม}ชิ้นนี้ \textbf{รังสรร}อย่างดี}\newline
        [\thaitext{\textbf{ประติมากรรม}ชิ้นนี้ \textbf{รังสรรค์}อย่างดี}]\newline
        (Eng: \textit{This sculpture is well-crafted})
    }
    \item{
        \thaitext{\textbf{สัปปะรส} กรอบ อร่อย}
        [\thaitext{\textbf{สับปะรด} กรอบ อร่อย}]\newline
        (Eng: \textit{Tasty pineapple})
    }
\end{itemize}
\end{flushleft}

\subsubsection{Typos}
Typos or typographical errors are unintended text usually caused by striking an incorrect key on a keyboard. It is mainly due to human errors. Although a spell checker has been developed on many platforms, many typos can still be found in the corpus. 

Typos can be classified into two classes; a non-word error and a real-word error. A non-word error is where a misspelt word conveys no meaning in the language; in the worse situation, a real-word error is a misspelt word that turns into a word that the writer does not mean to write \cite{kukich1992techniques}. Both cases can be easily detectable by the annotators if a word has a low edit distance from another word suited more to the context. However, it is accidental, so it presumably has no applicable semantics. 

\section{The Impact on Sentiment Analysis}
In this section, we propose two approaches to incorporate misspelling semantics into a sentiment classifier.  

\begin{figure}[!h]
    \begin{center}
        \includegraphics[width=0.45\textwidth]{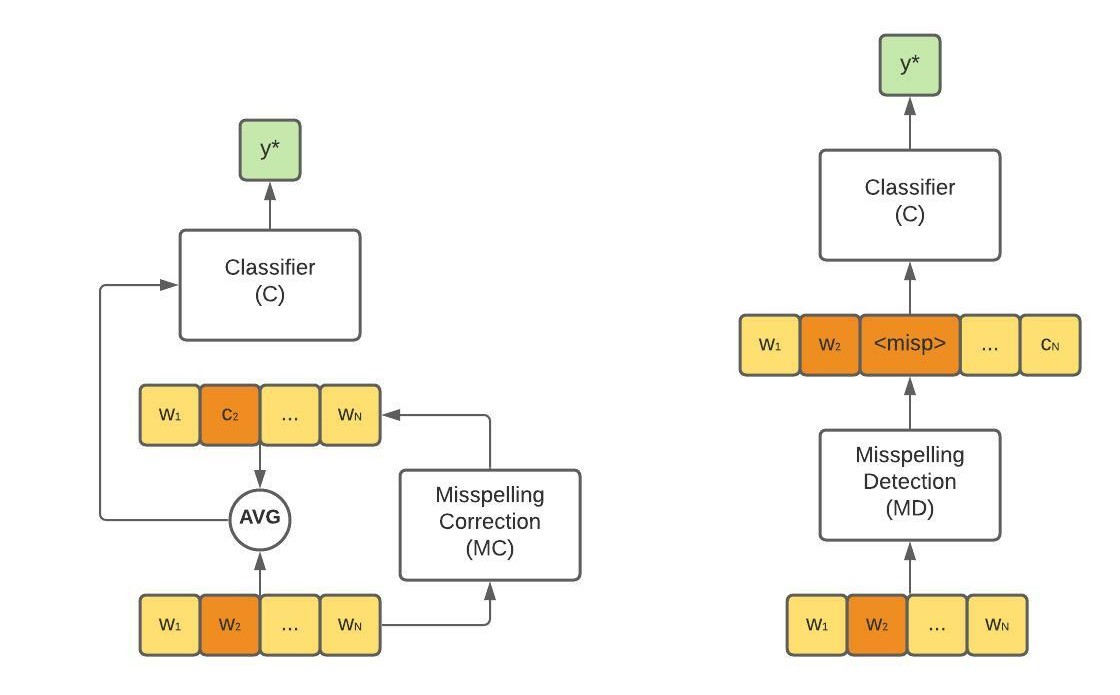}
        \caption{Overview architecture of Misspelling Average Embedding (left) and Misspelling Semantics Tokens (right)}
        \label{arch}
    \end{center}
\end{figure}

\subsection{Misspelling Average Embedding}

Misspelling Average Embedding (MAE) is based on the hypothesis that the embedding of a misspelt word and its correct word encode different semantics. Both embeddings could be complementary to each other. MAE uses the average of the embedding from the misspelt and its correct token as a representation of a word.

To formally define MAE, given a sentence $w = {w_1, w_2, w_3, … w_N}$ where $N$ is the total number of words and a misspelling correction model $MC(w_i) = c_i$, an embedding $E$ and a classifier $C$. The MAE computes a vector $w* = AVG(E(w),E(MC(w))$ where $AVG$ is an average function. MAE, then, uses $w*$ as an input to $C$ to get a prediction $y*$ (see Figure~\ref{arch}).

Conceptually, MAE can be applied both in training and testing time. However, we only presented results on the testing time. We expected that applying MAE during training could yield larger improvement, but we leave it for future study.

\subsection{Misspelling Semantic Tokens}
In Misspelling Semantic Tokens (MST), we introduce additional tokens to indicate the location of the misspelt words. We hypothesize that locating the misspelling is sufficient for a model to get a better language understanding. It requires only a misspelling detection which is significantly easier to build. However, it requires re-training.

There are four additional tokens introduced as misspelling semantic.
\textit{<int>} for intentionally misspellings, \textit{<msp>} for unintentionally misspellings, \textit{<lol>} for repeated `5` and \textit{<rep>} for other repeated characters. The last two were introduced because they have been studied and confirmed to have a close correlation with the sentiment. We differentiate repeated `5` with other repeated characters as it is the most common repeated character and always has its distinct meaning (it mimics hahahah sound in Thai). It could be more beneficial to a model to separately treat it from other types of misspelling.

Mathematically, given a sentence $w = {w_1, w_2, w_3, … w_N}$ where $N$ is the total number of words, an embedding $E$ and a classifier $C$ and a misspelling detection model $MD(w_i)$ defined as follows.

\resizebox{.9\hsize}{!}{
    $MD(w_i) = \begin{cases}
        \text{<lol>}, & \text{if $w_i$ has repeated 5}.\\
        \text{<rep>}, & \text{if $w_i$ has other repeated characters}.\\
        \text{<int>}, & \text{if $w_i$ is an intentionally misspelling.}.\\
        \text{<msp>}, & \text{if $w_i$ is an unintentionally misspelling}.\\
        Null, & \text{otherwise}.
    \end{cases}$
}

Firstly, we introduce 4 additional tokens to the embedding E with random weight initialisation. MST, then, transforms the sentence $w \in W$ into $ s* = {w_1, MD(w_1), w_2, MD(w_2), …, w_N, MD(w_N)}$. The $s*$ is used to re-train the embedding $E$ and the classifier $C$. Finally, use re-trained $E$ and $C$ to determine the prediction $y*$ (see Figure~\ref{arch}).

\section{Evaluation}
\subsection{Evaluation on non-contextual embedding}
\label{sec:non-contextual}
We applied MAE and MST on fastText embedding \cite{bojanowski2017fasttext}  in two settings; pre-trained embedding and embedding trained from scratch. For the former, we used pre-trained Thai fastText from \citeauthor{grave2018learningfasttext}. It was trained on Common Crawl and Wikipedia using CBOW with position-weights, in dimension 300, with character n-grams of length 5, a window of size 5 and 10 negatives. This setting represents a more practical situation where people can not access a large corpus but published models.

In the second setting, we used the VISTEC-TP-TH-2021 corpus \cite{limkonchotiwat-etal-2021-handling} to train another fastText model with the same settings (except using embedding dimension = 100). The corpus contains 49,997 sentences with 3.39M words from Twitter from 2017-2019. The misspellings and their corrected words were manually annotated by linguists. Misspellings due to Thai orthographic signs were discarded to align with our misspelling criteria. This represents a situation where a semi-large corpus is accessible.

We then trained a LSTM on top of these two embeddings, using Wisesight Train as input with batch size 256 in 100 epochs. The embeddings were frozen throughout the training step. The Wisesight Validation was used to select the best model. For misspelling correction (MC) and misspelling detection (MD), we used a dictionary-based model collected from our annotated corpus. This simulates a realistic situation in poorly-resourced languages where no accurate misspelling model is publicly available. 

\textbf{Results and Analysis}
We report micro F1 score in Table~\ref{tab:fasttext}. For the baseline where misspellings are kept intact without extra preprocessing (\textit{NONE}), we got 64.58\% and 66.68\% F1 from the pre-trained and from-scratch embeddings respectively. However, in another baseline where misspellings are normalised before training embedding (\textit{NORM}), the performance drops 0.6\% from the \textit{NONE} baseline. It suggests that misspelling normalisation on training data can be suboptimal.

Results from our MAE and MST methods
confirm our hypothesis. Both MAE and MST improved the F1 score by 0.4-1.95\%. MAE on pre-trained fastText gives only slight gains. The improvement is much clearer with our fastTest trained from scratch. The gains from MST suggest that locating the misspelling is also helpful; however, it is worth bearing in mind that MST requires re-training, and might not be suitable in many circumstances. 
When MAE and MST are applied together, we achieve the biggest boost, 1.95\% and 2.55\% over the \textit{NONE} and \textit{NORM} baselines. This confirms our hypothesis that misspelling has hidden semantics that are useful for sentiment-related tasks.

To further analyse, we report F1 on a subset of the test set where a sentence has at least one misspelt word. Even though normalised sentences were generally better than sentences with misspelling intact, MAE can boost the F1 to reach higher accuracy.

\begin{table}
    \centering
    \resizebox{\columnwidth}{!}{
        \begin{tabular}{|l|l|c|c|c|}
           
          \hline
          &
          Model &
          F1&
          F1 on misp&
          F1 on norm\\
          \hline
          \hline
          Baseline &
          Pre-trained fasttext (NONE) &
          64.58&
          58.41&
          59.09\\
          \hline
          Our &
          +MAE &
          \textbf{64.92}&
          59.43&
          -\\
          \hline
          \hline
          Baseline &
          VISTEC-TP-TH-2021 (NONE) &
          66.68&
          64.43&
          65.80\\
          \cline{2-5}
          &
          VISTEC-TP-TH-2021 (NORM) &
          66.08&
          61.93&
          62.95\\
          \hline
          Our &
          +MAE &
          66.90&
          65.11&
          -\\
          \cline{2-5}
          &
          +MST &
          68.06&
          62.84&
          63.30\\
          \cline{2-5}
          &
          +MAE+MST &
          \textbf{68.63}&
          63.98&
          -\\
          \hline
        \end{tabular}
    }
    \caption{Micro-F1 from LSTM classifer on top of a static embedding. It includes testing results from test data (F1), test data that has at least one misspelling word (F1 on misp) and its normalisation (F1 on norm).}
    \label{tab:fasttext}
\end{table}

\subsection{Evaluation on contextual embedding}
In contextual embedding setting, we experimented on a pre-trained Thai monolingual model, WangchanBERTa \cite{lowphansirikul2021wangchanberta}, a language model based on the RoBERTa-base architecture. It is a state-of-the-art model in Thai trained on a large corpus curated from diverse domains of social media posts, news articles and other publicly available datasets. 
The custom embedding layer was implemented on the output embedding for MAE.

We evaluated our approaches in two settings; a fully fine-tuned setting where the model was trained on the whole Wisesight training set and a few-shot setting where a model was trained by only 3000 training samples. Because MST introduces four additional tokens, we found that a longer training time was required to optimize the new token embeddings. So, throughout the experiment, the model was fine-tuned with batch size 32 in 10 epochs, using Wisesight validation to select the best model. However, because the training data in the few-shot setting was significantly less than the entire corpus, the training time was set to 40 epochs instead. Other parameters were set as default. The same MC and MD from the previous experiment were used.

To avoid mismatch tokenization between the normalised form and its misspelling, the first subtoken of the normalised form was duplicated to match the number of subtokens of the misspelling form. It is to guarantee that both embeddings can be averaged directly in MAE.

This experiment considers only \textit{NONE} baseline (pre-train/fine-tune with original text with misspelling unchanged) as it is more widely used in practice.

\textbf{Results and Analysis}
Results are shown in Table~\ref{tab:wangchanberta}, and convey a similar conclusion to the previous experiments. Overall, MAE and MST improve the F1 score by 0.2-0.37\%. The improvements are slightly less than in Section~\ref{sec:non-contextual}; this may be because the model has learnt the misspelling semantics during its pre-training. Further study on how a pre-trained language model handles misspelling is needed.

\begin{table}
    \centering
    \resizebox{\columnwidth}{!}{
        \begin{tabular}{|l|l|c|c|c|}
            
          \hline
          &
          Experiment &
          F1&
          F1 on misp&
          F1 on norm\\
          \hline
          \hline
          Few-shot &
          WangchanBERTa (NONE)
          &
          65.11&
          64.66&
          65.34\\
          \cline{2-5}
          &
          +MAE &
          65.41&
          65.57&
          -\\
          \cline{2-5}
          &
          +MST &
          \textbf{65.48}&
          65.68&
          64.77\\
          \cline{2-5}
          &
          +MAE+MST&
          65.33&
          65.23&
          -\\
          \hline
          \hline
          Fine-tuned &
          WangchanBERTa (NONE) &
          73.72&
          71.25&
          72.93\\
          \cline{2-5}
          &
          +MAE &
          73.57&
          70.80&
          -\\
          \cline{2-5}
          &
          +MST &
          \textbf{73.90}&
          72.39&
          71.82\\
          \cline{2-5}
          &
          +MAE+MST&
          73.68&
          71.70&
          -\\
          \hline
        \end{tabular}
    }
    \caption{Micro-F1 from WangchanBERTa on test data (F1), test data that has at least one misspelling word (F1 on misp) and its normalisation (F1 on norm). It includes results from few-shot setting (trained on 3000 training samples) and fine-tuned settings (trained on all training samples)}
    \label{tab:wangchanberta}
\end{table}

\section{Conclusion}

In this research, we introduce a new fine-grained annotated corpus of misspelling in Thai, including misspelling intention and its patterns. We highlight the semantics that can be exploited for language understanding tasks. Two approaches were demonstrated to incorporate the misspelling semantics for a sentiment analysis task. The experiments show that our approaches can improve existing models up to 2\%. They require only a simple dictionary-based misspelling detection and/or misspelling correction. However, our methods are less useful in pre-trained/fine-tuning settings with large language models. 

Overall, the experiments confirmed our hypothesis that misspellings contain hidden semantics which are useful for language understanding tasks while blindly normalising misspelling is harmful and suboptimal. Understanding misspelling semantics could support NLP researchers in devising better strategies to embrace unexpected content at either training or inference time.

\section*{Acknowledgements}

The authors acknowledge support from the UK EPSRC via the Sodestream project (Streamlining Social Decision-Making for Enhanced Internet Standards, grant EP/S033564/1), and from the Slovenian Research Agency for research core funding (No. P2-0103 and No. P5-0161).

\section{References}\label{reference}

\bibliographystyle{lrec}
\bibliography{references}

\end{document}